\documentclass[11pt]{article}
\usepackage[round]{natbib}
\usepackage[paper1]{paper_style/kirin-papers}
\usepackage{caption}
\titleformat{\section}{\sffamily\bfseries\color{KirinAccent}\fontsize{15.2pt}{17.5pt}\selectfont}{\thesection}{0.6em}{}
\titleformat{\subsection}{\sffamily\bfseries\color{KirinInk}\fontsize{12.2pt}{14.2pt}\selectfont}{\thesubsection}{0.55em}{}

\KirinSetShortTitle{Reading Human Preference}
\hypersetup{pdftitle={Reading Human Preference from Looped Transformer Internal States},pdfauthor={Jan Kirin}}
\begin{document}
\KirinTitleBlock
  {Preprint \textperiodcentered{} Version 3}
  {Reading Human Preference}
  {from Looped Transformer Internal States}
  {Jan Kirin}
  {\href{https://orcid.org/0009-0005-2681-9027}{ORCID 0009-0005-2681-9027}}
  {November 2026}
  {Versions~1 and~2 were titled \emph{Relational Preference Encoding in Looped Transformer Internal States}. This version replaces them with a corrected evaluation and withdraws their evidence that the model stores preference as a comparison between replies. Appendix~\ref{app:corrections} lists every changed headline claim.}

\begin{abstractbox}
Show a frozen language model two replies to the same request, one of which a person preferred. Can a small probe on the model's hidden states tell which? On Ouro-2.6B-Thinking, a model that runs its layers four times per token, and Anthropic's HH-RLHF data, the answer is yes, but only faintly. A linear probe picks the preferred reply 61.5\% of the time, a small nonlinear network that scores each reply separately 64.9\%, and the evaluator trained for version~1 reaches 64.7\% on the test split: clearly better than simple rules such as picking the longer reply, but weak. Whether preference is stored as a comparison between the two replies or as a property of each is still open. A linear probe cannot tell, and a network that reads both replies together has an estimated edge of only 0.7 points (95\% CI 0.0 to 1.4) over one that scores each reply separately. These numbers hold under three controls: scoring every pair in both orders, splitting the data by pair, and, for the probes, keeping the end of each conversation, where the two replies differ. Version~1's 95.2\% and 84.5\% were obtained without the first two and do not hold; version~2's corrected relational-probe accuracy, obtained without the third, was four to five points too low. This version replaces both.
\end{abstractbox}

\section{Introduction}

Think of a taste test: two cups, and a person who has said which one they liked better. Now replace the cups with two replies to the same request, and do not ask the language model anything. Just look at its hidden states as it reads each reply. Can a probe, a small classifier trained on those states, tell which reply the person preferred? If it can, preference is represented in a form that a small external reader can extract, without training the model as a reward model \citep{christiano2017deep,ouyang2022training}.

I ask this of Ouro-2.6B-Thinking \citep{zhu2025ouro}, a model that runs the same stack of layers four times for every token. The short answer is yes, but faintly. Where a coin would get five pairs in ten right, the readouts get about six.

\begin{figure}[t]
\centering
\includegraphics[width=0.92\linewidth]{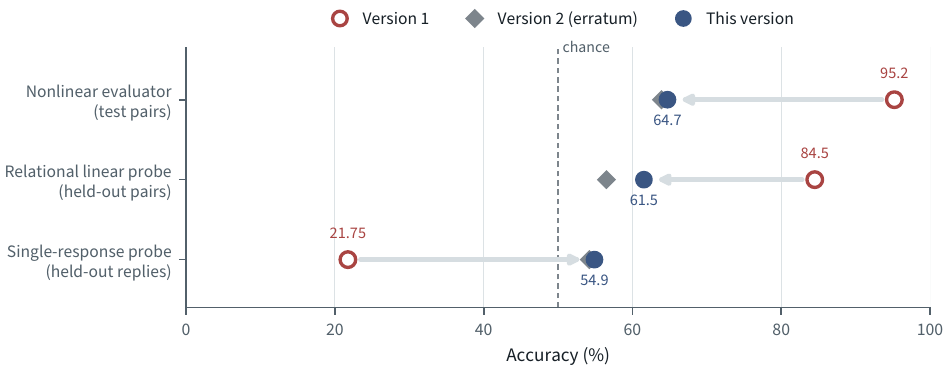}
\caption{\textbf{Version~1's headline numbers and the corrected values.} Accuracy at picking the human-preferred reply for the three readouts version~1 reported (open circles), version~2's corrections (diamonds), and the values here (filled circles). The evaluator is scored on the 8,552 HH-RLHF test pairs: by version~1 with the preferred reply always first, by version~2 on Ouro-RLTT at 768 tokens with ties counted as errors, and here on Ouro-Thinking at 1,024 tokens in both orders. The probes are scored on held-out training-split pairs, split by pair; the single-response probe one reply at a time. Version~2's probe values use all four loops, right truncation and ties counted as errors (all four loops here: 60.6\% and 54.3\%); version~1's relational probe and the probes here use the last loop; the evaluator reads all four loops. Two diamonds lie almost under the filled circles (63.9\% and 54.2\%). Table~\ref{tab:corrections} lists every changed number.}
\label{fig:summary}
\end{figure}

Version~1 of this paper \citep{kirin2026rpe} gave a much stronger answer: a small evaluator picked the preferred reply 95.2\% of the time, and a linear probe 84.5\%. It added that preference is encoded relationally, as a comparison between the two replies rather than as a property of each, because a probe scoring each reply on its own performed below chance (21.75\%). None of those numbers holds (Figure~\ref{fig:summary}): they were obtained without three controls described below. This version measures the same readouts again, on the same model and data, and finds the following.

\paragraph{Preference is readable, weakly} A linear probe on the hidden states picks the preferred reply in about 61\% of held-out pairs (Section~\ref{sec:probes}), and the nonlinear evaluator of version~1 in about 65\% (Section~\ref{sec:evaluator}). Both are above chance but far from perfect, and both beat simple rules such as picking the longer reply (Section~\ref{sec:surface}).

\paragraph{Linear probes cannot settle whether preference is relational} Is preference a property of each reply, like a price tag, or does it exist only in the comparison, like the winner in rock-paper-scissors, where no hand has a value of its own? Version~1 argued for the comparison. A linear probe cannot decide this, because scoring the difference of two replies is the same as reading a price tag off each reply and subtracting. A probe trained on differences and a probe trained on single replies are therefore the same kind of reader, trained in two ways, and given the same features they rank held-out pairs within two points of each other (Section~\ref{sec:probes}). Two small nonlinear networks can test the question. One that reads both replies together has a small estimated edge over one that scores each reply separately, but the test, fixed in advance, leaves it unresolved (Section~\ref{sec:relational}). Version~1's evidence for relational encoding does not survive the controls below.

\paragraph{The answer depends on three controls} A pairwise evaluation can flatter a readout in ways that standard diagnostics do not catch. Version~1's evaluator, tested with the preferred reply always shown first, had learned to prefer the first reply, like a taster with a habit of reaching for the left cup; scoring both orders removes this (Section~\ref{sec:position}). Its probes were tested on rows derived from their own training pairs (for the relational probe, an exam set from the homework); splitting by pair removes this (Section~\ref{sec:leakage}). In version~2's corrected probes, truncating long conversations from the right removed the final replies, the only part where the two members of a pair differ, so that 7.7\% of held-out pairs became identical and were scored as errors; truncating from the left removes this (Section~\ref{sec:truncation}). The first two were missing in version~1, the third in version~2's corrected probes. Each has a cheap test.

\paragraph{Caveats}
\begin{itemize}[leftmargin=*,itemsep=1pt]
\item One model family (Ouro-2.6B and two of its fine-tuned versions) and one dataset.
\item The probes read hidden states averaged over the last 384 tokens; longer conversations lose their beginning.
\item The evaluator is version~1's trained checkpoints, re-evaluated; I did not retrain it.
\item Some analyses were designed after the problems with versions 1 and~2 were found; Appendix~\ref{app:repro} says which.
\end{itemize}

\section{Setup}\label{sec:setup}

\paragraph{Model} Ouro-2.6B-Thinking \citep{zhu2025ouro} applies the same 48 transformer layers four times per token; I call each pass a loop. I read the hidden state after each loop (after the final normalization) and average it over the tokens of the input, which gives four 2,048-dimensional vectors per response. For comparison I also read the base Ouro-2.6B and Ouro-RLTT \citep{williams2026rltt}, a version of Ouro fine-tuned with reinforcement learning. All models are frozen and run in bfloat16.

\paragraph{Data} HH-RLHF \citep{bai2022training} contains conversations in which a person chose one of two final assistant replies; the two members of a pair share everything except that reply. The probes use 40,000 pairs from the training split, divided by pair into 32,000 for training and 8,000 for evaluation, so that no pair contributes to both. The evaluator was trained in version~1 on 50,000 training-split pairs and is tested here on all 8,552 pairs of the test split.

\paragraph{Probes} I use two linear probes. The \emph{relational probe} is a logistic regression on the difference $h_{\text{chosen}} - h_{\text{rejected}}$ with no intercept, so that swapping the two replies exactly negates its score. The \emph{single-response probe} is a logistic regression with an intercept that labels each reply as chosen or rejected on its own; to compare it with the relational probe I use its two scores to rank the pair, where the intercept cancels.

One fact about these two probes matters for everything that follows. The relational probe's score is $w^\top h_{\text{chosen}} - w^\top h_{\text{rejected}}$: a score of each reply, computed twice and subtracted. The two probes therefore rank pairs with the same family of functions and differ only in how they are trained.

Both use an L2 penalty of $\tfrac12\lVert w\rVert^2/n$, where $n$ is the number of training rows (32,000 pairs for the relational probe, 64,000 replies for the single-response probe, so the relational probe is regularized twice as strongly), and are fit to convergence with Newton's method in double precision on raw features, either the last loop (2,048 dimensions) or all four loops concatenated (8,192). Inputs keep the last 384 tokens of each conversation (Section~\ref{sec:truncation}).

\paragraph{Evaluator} Version~1's evaluator pools each loop's token states with a learned low-rank attention, takes the difference between the two replies, normalizes it, runs a two-layer GRU over the four loops, and maps the result to a score; it has 4.7M parameters and was trained for five epochs with the order of the two replies swapped at random. I score every pair in both orders and count a pair as correct when the preferred reply wins on average over the two orders, $s(c, r) - s(r, c) > 0$.

\paragraph{Statistics} Accuracies are over pairs, in percent, with exact ties counted as one half; differences between accuracies are in percentage points. Intervals are 95\% bootstrap intervals over pairs (10,000 draws), shown in square brackets. Two readouts measured on the same pairs are compared by bootstrapping their per-pair difference.

\section{What can be read}\label{sec:results}

\subsection{A linear probe picks the preferred reply about 61\% of the time}\label{sec:probes}

\begin{figure}[t]
\centering
\includegraphics[width=\linewidth]{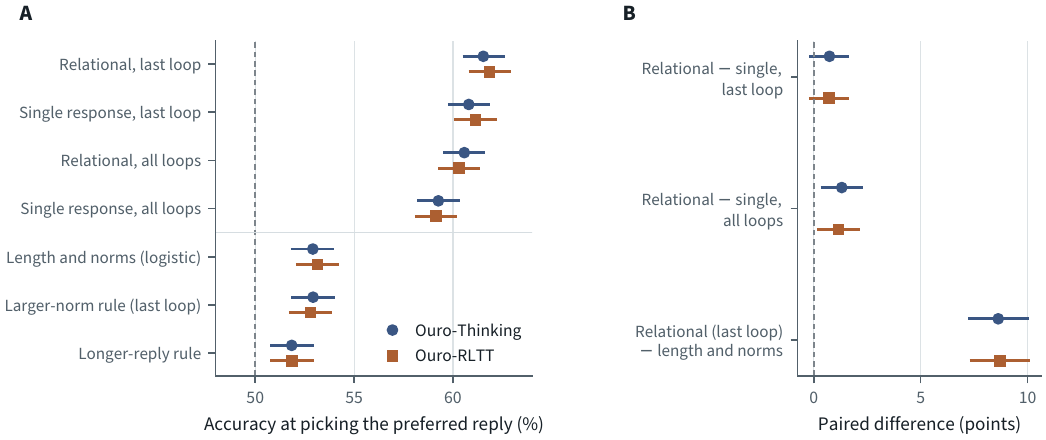}
\caption{\textbf{What the probes read} (8,000 held-out HH-RLHF pairs, last 384 tokens, both backbones). (A)~Accuracy at picking the preferred reply, with 95\% bootstrap intervals and ties counted as one half. The single-response probe scores each reply on its own and ranks the two. Below the line: the longer-reply rule, the larger-norm rule (last-loop norm), and a logistic regression on the differences in length and in the norm of each loop's state. (B)~Paired differences on the same pairs, with 95\% intervals.}
\label{fig:probes}
\end{figure}

Table~\ref{tab:probes} and Figure~\ref{fig:probes} give the probes on the 8,000 held-out pairs. On the last loop, the relational probe picks the preferred reply in 61.5\% of pairs and the single-response probe, ranking the same pairs, in 60.8\%; the paired difference, $+0.7$ points, is not distinguishable from zero. Concatenating all four loops does not help. The relational probe reads 1.0 point less ($[-0.04, 2.0]$, an interval that includes zero), and the single-response probe loses more, which widens the gap between them to 1.3 points ($[0.3, 2.3]$). Read one reply at a time, the single-response probe labels 54.9\% of replies correctly, above chance; version~1's below-chance 21.75\% was an artifact (Section~\ref{sec:leakage}).

Version~1 concluded that preference is ``predominantly relational'' because its single-response probe scored far below its relational one. With both probes trained on the same features and compared on the same pairs, the estimated gap is below two points, with the all-loop interval reaching 2.3, and on the last loop, where both probes do best, it is not distinguishable from zero.

A linear probe cannot settle the question in either direction. The relational probe's score is a score of each reply, $w^\top h$, computed twice and subtracted, which is exactly how the single-response probe ranks a pair. The two differ only in training: a pairwise loss on 32,000 pairs against a per-reply loss on 64,000 replies, with different regularization. What the comparison shows is that the two ways of training give similar rankings, not that preference is a property of each reply alone. Each reply's input also includes the conversation before it, so neither probe reads a reply out of context, and because the states are averaged over inputs of different lengths, the shared conversation does not cancel exactly in the difference.

\begin{table}[t]
\centering
\caption{Linear probes and simple rules on 8,000 held-out HH-RLHF pairs (Ouro-2.6B-Thinking, last 384 tokens). Accuracy (\%) at picking the preferred reply, with 95\% bootstrap intervals; the single-response probe ranks the pair by its two scores. Both probes rank with a score per reply and differ only in training (Section~\ref{sec:probes}). Differences are in points, paired over the same pairs. The simple rules take their direction from the training pairs (Section~\ref{sec:surface}).}
\label{tab:probes}
\begin{tabular}{@{}lll@{}}
\toprule
Probe & Last loop & All four loops \\
\midrule
Relational (difference of the two replies) & 61.5 $[60.5, 62.6]$ & 60.6 $[59.5, 61.6]$ \\
Single response, used to rank the pair & 60.8 $[59.7, 61.9]$ & 59.3 $[58.2, 60.3]$ \\
Relational minus single response & $+0.7$ $[-0.2, +1.7]$ & $+1.3$ $[+0.3, +2.3]$ \\
\addlinespace
\multicolumn{3}{@{}l}{\emph{Simple rules}} \\
Longer reply & \multicolumn{2}{c}{51.8 $[50.8, 53.0]$} \\
Larger last-loop norm & \multicolumn{2}{c}{52.9 $[51.8, 54.0]$} \\
Logistic regression on length and norm differences & \multicolumn{2}{c}{52.9 $[51.8, 54.0]$} \\
\bottomrule
\end{tabular}
\end{table}

\paragraph{Backbones} Ouro-RLTT gives the same picture as Thinking (Figure~\ref{fig:probes}): its relational probe reads 61.9\% ($[60.8, 62.9]$) on the last loop and its single-response probe 61.1\% when ranking the same pairs. The gap between them is $+0.7$ points ($[-0.2, +1.7]$) on the last loop and $+1.2$ points ($[+0.2, +2.2]$) with all four loops, and both probes beat the length-and-norm regression (53.2\%) by about eight points. Under version~2's protocol (all four loops, right truncation, ties counted as errors), the relational probe read 55.5\% on the base model, 56.5\% on Thinking and 57.0\% on RLTT; only RLTT minus base was distinguishable from zero ($+1.5$ points, $[+0.4, +2.5]$, paired over the same pairs).

\subsection{A nonlinear evaluator reads somewhat more}\label{sec:evaluator}

Scored in both orders on all 8,552 test pairs, version~1's evaluator (epoch~2, the checkpoint it reported) picks the preferred reply in 64.7\% of pairs ($[63.7, 65.7]$) on Ouro-Thinking, the model it was trained on, with its original 1,024-token inputs, and in 65.0\% ($[64.0, 66.0]$) on Ouro-RLTT with 768-token inputs; the backbone makes little difference. The other four epochs give 63.7\% to 65.3\% on Thinking (Table~\ref{tab:epochs}) and 64.2\% to 65.5\% on RLTT. Its accuracy is about three points above the linear probes', but the two differ in more than architecture: in the pairs they are scored on (the test split for the evaluator, held-out training-split pairs for the probes), the token limit and truncation side, and the training data. The gap cannot be attributed to the evaluator's learned pooling or to the way it combines the four loops. Section~\ref{sec:relational} tests nonlinear readouts on the probes' own pairs and split.

\subsection{The readouts beat simple length and norm rules}\label{sec:surface}

Perhaps the readouts track something simpler than preference, such as reply length. If people preferred longer replies, a probe could read length from the hidden states and look as if it read preference. That cannot be the whole story: simple rules do only slightly better than chance. On the probes' held-out pairs, picking the longer reply, picking the reply whose last-loop state has the larger norm, and a logistic regression on the differences in length and in each loop's norm reach 51.8\% to 52.9\% (Table~\ref{tab:probes}); the relational probe beats the regression by 8.6 points ($[7.2, 10.1]$). On the evaluator's test pairs, the longer reply (counted in tokens, at the 768-token limit of version~2's audit) is the preferred one in 52.9\% of pairs ($[51.9, 53.9]$) and the larger-norm reply in 52.3\%; the evaluator beats these rules by about 12 points (12.4 points, $[11.0, 13.8]$, over the norm rule). Version~1, looking at 500 pairs, had found rejected replies longer on average and the larger-norm reply preferred in only 43\% of them, and concluded that neither length nor norm separated the classes; on the full test split both lean slightly the other way. These comparisons show that the readouts beat the simplest shortcuts. They do not measure how much the readouts rely on length in other ways, which would need analyses matched or conditioned on length.

\subsection{A network that reads both replies together has a small, unresolved edge}\label{sec:relational}

A linear probe cannot tell whether preference is relational (Section~\ref{sec:probes}), but nonlinear readouts can test whether reading the two replies together helps. I trained two small networks on the probes' pairs and split, each reading the top 512 principal components of the same hidden states (Appendix~\ref{app:repro}). The per-reply network scores each reply in its conversation context and subtracts, $f(h_A) - f(h_B)$. The pair network, $g$, sees both replies at once, through their two states, their difference and their elementwise product, and scores the pair as $\tfrac12[g(A,B) - g(B,A)]$: swapping the replies still flips the sign of the score, but the score can depend on how the two replies relate. If the relation between the replies carries information that these networks can use, the pair network should come out ahead. Both networks chose their size and penalty from the same grid on a validation slice of the training pairs, and each network's score is the average of five seeds.

After the linear results were known, and before the first fit, I fixed the protocol and how its result would be judged. The pair network's lead counts as an advantage if its 95\% interval lies above zero, as no advantage above one point if the interval lies below one point, and as unresolved otherwise.

On the last loop of Ouro-Thinking, the per-reply network picks the preferred reply in 64.9\% of held-out pairs ($[63.8, 65.9]$) and the pair network in 65.6\% ($[64.5, 66.6]$). The pair network's lead, $+0.7$ points ($[0.0, 1.4]$), is unresolved: it shows neither a real advantage nor that any advantage is below one point. The two secondary comparisons, with all four loops and on Ouro-RLTT, give $-0.3$ ($[-1.0, 0.4]$) and $+0.0$ ($[-0.7, 0.7]$); both meet the rule for no advantage above one point. For these networks, then, scoring each reply separately reaches similar accuracy to reading the pair. This does not measure how much of the preference information in the states can be read from each reply separately. A pair network cut down to the per-reply network's size, a check the protocol lists, leads it by 0.65 points ($[-0.05, 1.36]$). The comparison is also narrow: in the primary analysis both networks chose the smallest width and the strongest penalty in the grid, and in all three analyses every chosen penalty lies at one end of it.

The per-reply network does beat the linear probe, by 3.3 points ($[2.3, 4.4]$) on the same pairs. About one point of that comes from its input: a linear probe on the same 512 components reads 62.5\%, and the per-reply network beats it by 2.3 points ($[1.5, 3.1]$). Scoring each reply nonlinearly therefore improves on the linear probe without reading the two replies together. The per-reply network's accuracy is close to the evaluator's 64.7\%, but the evaluator is scored on different pairs and differs in inputs and training, so this comparison does not show what gives the evaluator its edge.

\section{Three controls the measurement depends on}\label{sec:errors}

The numbers of Section~\ref{sec:results} hold under three controls; the third applies to the probes, and the evaluator keeps its own truncation (Section~\ref{sec:truncation}). Without each of them, version~1 or~2 obtained a number that looked plausible and passed the checks it ran. The first two moved version~1's numbers by 23 to 33 points; the third lowered version~2's relational probe by four to five points. For each I describe what the uncontrolled number looked like, why it arises, and the check that catches it; each subsection ends with the control in brief.

\subsection{Position bias: the evaluator learned to prefer the first reply}\label{sec:position}

\begin{figure}[t]
\centering
\includegraphics[width=\linewidth]{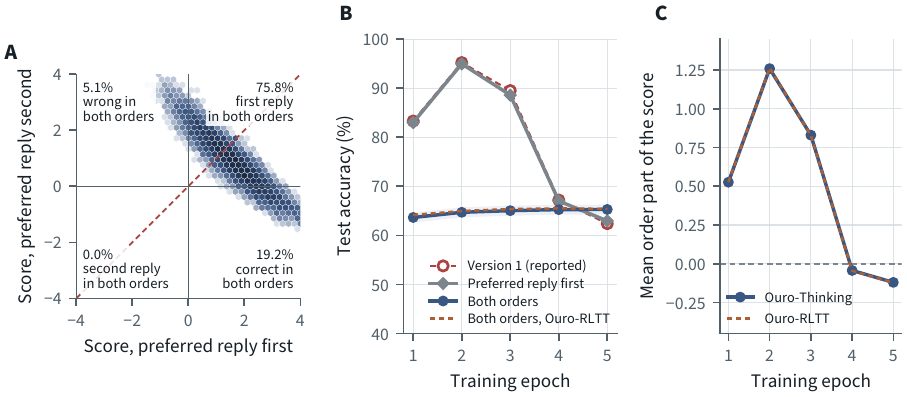}
\caption{\textbf{Version~1's evaluator learned the order of the replies} (8,552 HH-RLHF test pairs). (A)~For the epoch-2 checkpoint that version~1 reported, each pair's score with the preferred reply shown first against its score with the preferred reply shown second (Ouro-Thinking; shading is the log count; values beyond $\pm 4$, 0.5\% of pairs, are drawn at the edge). A positive score picks the first reply. Pairs below the dashed diagonal are correct when the two orders are averaged. (B)~Test accuracy after each training epoch: as version~1 reported it, under its protocol rerun here (preferred reply first), and averaged over both orders, with the 95\% interval shaded. (C)~The order offset, the mean order part of the score, $\tfrac12[s(c,r)+s(r,c)]$; the fixed-order accuracy rises and falls with it. In (B) and~(C), solid lines are Ouro-Thinking at 1,024 tokens, and the dashed line without markers is Ouro-RLTT at 768 tokens, which lies almost on top of them; in (B), the dashed line with open circles is version~1's report.}
\label{fig:position}
\end{figure}

Version~1 trained the evaluator with the two replies in random order but tested it with the preferred reply always first, counting a positive score as correct. On the full test split, with Ouro-Thinking, that protocol gives 94.9\%. Swapping the two replies drops it to 19.2\%. In 75.8\% of pairs the evaluator prefers whichever reply comes first in both orders (Figure~\ref{fig:position}A), so most of the 94.9\% measured the order, not the content. It is the taster with a habit of reaching for the left cup: put the preferred cup on the left every time, and they score almost perfectly while barely tasting. Splitting each score into a content part, $\tfrac12[s(c,r) - s(r,c)]$, which does not depend on the order, and an order part, $\tfrac12[s(c,r) + s(r,c)]$, shows why: the order part is on average 1.5 times as large as the content part, and its mean, the order offset, is $+1.26$. The content part alone gives the 64.7\% of Section~\ref{sec:evaluator}; I call its accuracy the order-independent accuracy.

Version~1 did run a check for order: it swapped the replies on 100 pairs and found that the scores moved in opposite directions (correlation $-0.94$; $-0.93$ on the full test split), and took this as evidence of a well-behaved evaluator. That correlation measures whether the score responds to the content; it says nothing about a constant offset added to every score, which is what inflated the accuracy. The check that catches it is to report the order-independent accuracy.

The same offset explains version~1's account of training. Version~1 reported that test accuracy rose from 83.3\% after epoch~1 to 95.2\% after epoch~2 and then fell to 62.4\% by epoch~5, while the training accuracy, measured with random order, rose steadily from 60\% to 70\%; it explained the peak by a learning-rate schedule that happened to stop training at the right time. Table~\ref{tab:epochs} and Figure~\ref{fig:position}B--C score all five checkpoints both ways. The fixed-order protocol reproduces version~1's curve to within a point. The order-independent accuracy does not peak at all: it rises from 63.7\% to 65.3\% and is flat from epoch~3 on. What rose and fell was the order offset, which is largest at epoch~2 and has vanished by epoch~4. The ``collapse'' was the offset disappearing; the learning-rate account explains an artifact. Version~1's training accuracy, measured with each example's order randomized, is yet another quantity: random order balances the offset across examples but does not remove it from each score. On the test pairs, random-order accuracy would have been 61.4\%, 57.1\%, 61.1\%, 65.1\% and 65.1\% after epochs 1 to~5, below the order-independent values while the offset was large; on the training data it rose from 60\% to 70\%.

\begin{scopebox}[Position bias in brief]
\textbf{Without it.} 94.9\% with the preferred reply shown first, 19.2\% with it shown second.\par
\textbf{Why.} A constant offset in the score favors whichever reply comes first.\par
\textbf{Control.} Score every pair in both orders and report the order-independent accuracy.
\end{scopebox}

\begin{table}[t]
\centering
\caption{Version~1's evaluator after each training epoch, on all 8,552 HH-RLHF test pairs (Ouro-2.6B-Thinking, 1,024-token inputs). Fixed order counts a pair as correct when the preferred reply, shown first, scores above zero, as version~1 did. Order-independent accuracy counts it as correct when the preferred reply wins on average over both orders. The order offset is the mean of $\tfrac12[s(c,r)+s(r,c)]$.}
\label{tab:epochs}
\begin{tabular}{@{}lrrrr@{}}
\toprule
Epoch & Version~1 reported & Fixed order & Order-independent [95\% CI] & Order offset \\
\midrule
1 & 83.3\% & 83.0\% & 63.7\% $[62.6, 64.7]$ & $+0.53$ \\
2 & 95.2\% & 94.9\% & 64.7\% $[63.7, 65.7]$ & $+1.26$ \\
3 & 89.5\% & 88.6\% & 65.0\% $[64.0, 66.1]$ & $+0.83$ \\
4 & 67.2\% & 67.0\% & 65.3\% $[64.2, 66.3]$ & $-0.04$ \\
5 & 62.4\% & 63.0\% & 65.3\% $[64.3, 66.3]$ & $-0.12$ \\
\bottomrule
\end{tabular}
\end{table}

\subsection{Group leakage: probes tested on their own training pairs}\label{sec:leakage}

Both of version~1's probes were built from rows derived from pairs, and both were split by row rather than by pair. The relational probe's data had two rows per pair, $+\Delta$ and $-\Delta$, so the test set contained the exact negation of training rows: an exam whose questions were the homework, read in a mirror. Because the probe is antisymmetric by construction, a probe that has fit $+\Delta$ scores $-\Delta$ correctly for free: the property version~1 cited as a safeguard made the leak complete. That probe reported 84.5\% on the last loop. Split by pair, on all four loops, it reads 56.5\% under the old truncation and 60.6\% under the corrected one (61.5\% on the last loop). The single-response probe's chosen and rejected rows were split independently, so 74\% of test replies had their pair partner in training. The probe picked up features specific to each pair, so a test reply whose partner had been in training, sharing its conversation but carrying the opposite label, tended to receive the partner's label. The probe scored 21.75\%, below chance. Split by pair it is above chance (Section~\ref{sec:probes}).

The check is to split by the unit that generates the rows, here the pair, and to count the units that cross the split; the count must be zero \citep{kapoor2023leakage}. Note that the two leaks announced themselves differently: the first as a strong result, the second as an interestingly bad one, which version~1 published as its key diagnostic.

\begin{scopebox}[Group leakage in brief]
\textbf{Without it, in version~1.} One probe far too good (84.5\%, on 1,000 examples), the other below chance (21.75\%).\par
\textbf{Why.} Rows built from the same pair sat on both sides of a split made by row.\par
\textbf{Control.} Split by pair, and count the pairs that cross the split; the count must be zero.
\end{scopebox}

\subsection{Truncation: the differing replies were cut off}\label{sec:truncation}

\begin{figure}[t]
\centering
\includegraphics[width=0.66\linewidth]{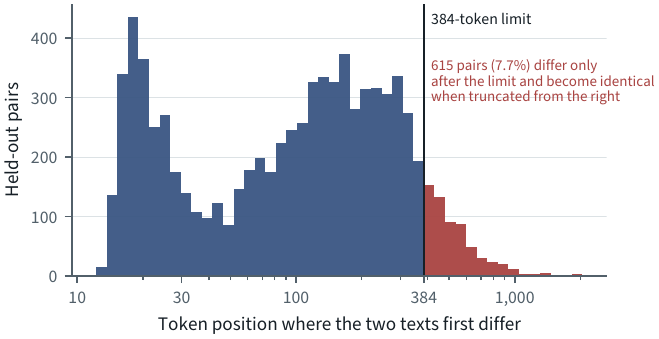}
\caption{\textbf{Truncating from the right removes the part of a pair that differs} (8,000 held-out training-split pairs). The token at which the preferred and the rejected text first differ. With the 384-token limit applied from the right, the 615 pairs that differ only after the limit (red) become identical inputs. A further 34 pairs have identical texts and are not shown.}
\label{fig:truncation}
\end{figure}

The two members of an HH-RLHF pair differ only in the final reply, at the end of the text. The tokenizer truncates long inputs from the right, so for long conversations it removes exactly the part that differs. It is like judging two essays that differ only in their last paragraph after tearing the last page off both. At the probes' 384-token limit, 649 of the 8,000 held-out pairs (8.1\%) were token-for-token identical: 615 because their two replies differ only after the limit (Figure~\ref{fig:truncation}), and 34 because their two texts are the same. Because both members of a pair were run in the same batch, their hidden states were identical too, the difference was exactly zero, and the probe's score was exactly zero; the rule ``correct if the score is positive'' counted every one of these pairs as an error. I confirmed this for all 649 pairs. The relational probe that version~2 reported at 56.5\% is therefore 61.5\% on the 7,351 pairs whose inputs differ. Giving the 649 tied pairs half credit instead raises it to 60.6\%, the same rounded accuracy that the left-truncated probe reaches on all four loops. Half credit alone thus accounts for the aggregate increase in this comparison; restoring the differing text can still change individual predictions. In version~2's audit of the evaluator, at 768 tokens, 125 test pairs were identical; its accuracy on the rest is 64.6\% rather than 63.9\%. I found no record of how version~1's probes truncated their inputs.

The check is to keep the end of the input (truncate from the left) when the classes differ at the end, to count identical inputs, and to score exact ties as one half. All probe results in Section~\ref{sec:results} do this; with the end kept, 34 of the 8,000 held-out pairs remain identical, all of them pairs whose two texts are the same. The evaluator results keep the evaluator's own right truncation. At 1,024 tokens it leaves 68 test pairs token-identical; excluding them raises the epoch-2 accuracy by 0.09 points, and by at most 0.12 over the five epochs.

\begin{scopebox}[Truncation in brief]
\textbf{Without it.} 649 of the 8,000 held-out pairs (8.1\%) had identical inputs, and each was scored as an error.\par
\textbf{Why.} For 615 of them, truncating from the right removed the final replies, the only text in which a pair differs.\par
\textbf{Control.} Keep the end of the input, count identical inputs, and score exact ties as one half.
\end{scopebox}

\section{Related work}\label{sec:related}

\paragraph{Probing} Linear probes test whether information is linearly decodable from a model's representations \citep{alain2016probes,belinkov2022probing}; decodability does not imply that the model uses the information, and probes need baselines and controls to be interpreted \citep{hewitt2019control}. Linear structure in representations has been found for many concepts, including truth \citep{marks2023geometry} and refusal \citep{arditi2024refusal}, and proposed as a general hypothesis \citep{park2023linear}. The surface rules of Section~\ref{sec:surface} are comparison baselines, not the control tasks of \citet{hewitt2019control}.

\paragraph{Preference and reward models} Reward models trained on human comparisons are the basis of RLHF \citep{christiano2017deep,ouyang2022training,bai2022training}. For language models they are usually trained end to end from a pretrained model \citep{ouyang2022training,bai2022training}; the readouts here train only a small head on frozen states, so they measure what the states already contain rather than what can be trained into them.

\paragraph{Evaluation pitfalls} Language models used as judges prefer responses by position \citep{zheng2023judge,wang2023fair}; the evaluator of Section~\ref{sec:position} is a small learned judge with the same bias. Leakage between training and test data through shared source items is a common cause of inflated results in machine-learning-based science \citep{kapoor2023leakage}.

\paragraph{Looped models} Looped and recurrent-depth transformers reuse their layers several times per token \citep{dehghani2019ut,geiping2025huginn,zhu2025ouro}. A companion paper probes the same Ouro models for whether an answer will be correct \citep{kirin2026opi}. Version~1 stated that, to its knowledge, Ouro was the only publicly released looped model with open weights and accessible per-loop states; Huginn \citep{geiping2025huginn} is another.

\section{Limitations and conclusion}

\paragraph{Limitations} All results come from Ouro-2.6B and two fine-tuned versions of it, on one dataset. The probes read states averaged over the last 384 tokens, which loses the beginning of long conversations and any information carried by token position. The evaluator was trained once, in version~1, on Ouro-Thinking features; I re-evaluated its saved checkpoints but did not retrain it, so its training choices (a 1,024-token limit with right truncation, random order swaps, five epochs) are its own. The nonlinear test of Section~\ref{sec:relational} covers one family of small networks, whose selected settings lie at the edge of its grid. HH-RLHF labels are noisy, and these are small readouts of a frozen model, so the numbers describe what its states make easy to read rather than the best accuracy achievable on this data.

\paragraph{Conclusion} The hidden states of a frozen looped language model carry human preference, but weakly: a linear probe reads it in about 61\% of pairs, and both version~1's evaluator and a small network that scores each reply separately in about 65\%. A network that reads both replies together has a small estimated edge in the primary comparison, which remains unresolved, and neither secondary comparison supports an edge above one point for the tested networks. These results show what the tested readers can extract, without determining how preference is represented in general. These numbers depend on three controls. Version~1 obtained its large numbers without the first two, and they do not hold; without the third, version~2's corrected relational probe came out too low; the standard diagnostics caught none of this. For any evaluation built on pairs I would now score both orders, split by pair and count the pairs that cross the split, and check how many inputs truncation makes identical.

\bibliographystyle{plainnat}
\bibliography{refs}

\begin{thebibliography}{18}
\providecommand{\natexlab}[1]{#1}
\providecommand{\url}[1]{\texttt{#1}}
\expandafter\ifx\csname urlstyle\endcsname\relax
  \providecommand{\doi}[1]{doi: #1}\else
  \providecommand{\doi}{doi: \begingroup \urlstyle{rm}\Url}\fi

\bibitem[Alain and Bengio(2016)]{alain2016probes}
Guillaume Alain and Yoshua Bengio.
\newblock Understanding intermediate layers using linear classifier probes.
\newblock \emph{arXiv preprint arXiv:1610.01644}, 2016.

\bibitem[Arditi et~al.(2024)Arditi, Obeso, Syed, Paleka, Panickssery, Gurnee,
  and Nanda]{arditi2024refusal}
Andy Arditi, Oscar Obeso, Aaquib Syed, Daniel Paleka, Nina Panickssery, Wes
  Gurnee, and Neel Nanda.
\newblock Refusal in language models is mediated by a single direction.
\newblock \emph{arXiv preprint arXiv:2406.11717}, 2024.

\bibitem[Bai et~al.(2022)Bai, Jones, Ndousse, et~al.]{bai2022training}
Yuntao Bai, Andy Jones, Kamal Ndousse, et~al.
\newblock Training a helpful and harmless assistant with reinforcement learning
  from human feedback.
\newblock \emph{arXiv preprint arXiv:2204.05862}, 2022.

\bibitem[Belinkov(2022)]{belinkov2022probing}
Yonatan Belinkov.
\newblock Probing classifiers: Promises, shortcomings, and advances.
\newblock \emph{Computational Linguistics}, 48\penalty0 (1):\penalty0 207--219,
  2022.
\newblock arXiv:2102.12452.

\bibitem[Christiano et~al.(2017)Christiano, Leike, Brown, Martic, Legg, and
  Amodei]{christiano2017deep}
Paul~F. Christiano, Jan Leike, Tom Brown, Miljan Martic, Shane Legg, and Dario
  Amodei.
\newblock Deep reinforcement learning from human preferences.
\newblock In \emph{Advances in Neural Information Processing Systems}, 2017.
\newblock arXiv:1706.03741.

\bibitem[Dehghani et~al.(2019)Dehghani, Gouws, Vinyals, Uszkoreit, and
  Kaiser]{dehghani2019ut}
Mostafa Dehghani, Stephan Gouws, Oriol Vinyals, Jakob Uszkoreit, and {\L}ukasz
  Kaiser.
\newblock Universal transformers.
\newblock In \emph{International Conference on Learning Representations}, 2019.
\newblock arXiv:1807.03819.

\bibitem[Geiping et~al.(2025)Geiping, McLeish, Jain, Kirchenbauer, Singh,
  Bartoldson, Kailkhura, Bhatele, and Goldstein]{geiping2025huginn}
Jonas Geiping, Sean McLeish, Neel Jain, John Kirchenbauer, Siddharth Singh,
  Brian~R. Bartoldson, Bhavya Kailkhura, Abhinav Bhatele, and Tom Goldstein.
\newblock Scaling up test-time compute with latent reasoning: A recurrent depth
  approach.
\newblock \emph{arXiv preprint arXiv:2502.05171}, 2025.

\bibitem[Hewitt and Liang(2019)]{hewitt2019control}
John Hewitt and Percy Liang.
\newblock Designing and interpreting probes with control tasks.
\newblock In \emph{Proceedings of EMNLP-IJCNLP}, 2019.
\newblock arXiv:1909.03368.

\bibitem[Kapoor and Narayanan(2023)]{kapoor2023leakage}
Sayash Kapoor and Arvind Narayanan.
\newblock Leakage and the reproducibility crisis in machine-learning-based
  science.
\newblock \emph{Patterns}, 4\penalty0 (9), 2023.
\newblock arXiv:2207.07048.

\bibitem[Kirin(2026{\natexlab{a}})]{kirin2026opi}
Jan Kirin.
\newblock Operational proto-introspection in looped language models:
  Process-quality taps, executable branching, and the readout--control
  boundary.
\newblock \emph{arXiv preprint arXiv:2607.18553}, 2026{\natexlab{a}}.
\newblock Version 3.

\bibitem[Kirin(2026{\natexlab{b}})]{kirin2026rpe}
Jan Kirin.
\newblock Relational preference encoding in looped transformer internal states.
\newblock \emph{arXiv preprint arXiv:2604.09870v1}, 2026{\natexlab{b}}.
\newblock Version 1; version 2 (arXiv:2604.09870v2) carries the erratum.

\bibitem[Marks and Tegmark(2023)]{marks2023geometry}
Samuel Marks and Max Tegmark.
\newblock The geometry of truth: Emergent linear structure in large language
  model representations of true/false datasets.
\newblock \emph{arXiv preprint arXiv:2310.06824}, 2023.

\bibitem[Ouyang et~al.(2022)Ouyang, Wu, Jiang, et~al.]{ouyang2022training}
Long Ouyang, Jeff Wu, Xu~Jiang, et~al.
\newblock Training language models to follow instructions with human feedback.
\newblock In \emph{Advances in Neural Information Processing Systems}, 2022.
\newblock arXiv:2203.02155.

\bibitem[Park et~al.(2024)Park, Choe, and Veitch]{park2023linear}
Kiho Park, Yo~Joong Choe, and Victor Veitch.
\newblock The linear representation hypothesis and the geometry of large
  language models.
\newblock In \emph{International Conference on Machine Learning}, 2024.
\newblock arXiv:2311.03658.

\bibitem[Wang et~al.(2024)Wang, Li, Chen, et~al.]{wang2023fair}
Peiyi Wang, Lei Li, Liang Chen, et~al.
\newblock Large language models are not fair evaluators.
\newblock In \emph{Proceedings of ACL}, 2024.
\newblock arXiv:2305.17926.

\bibitem[Williams and Tureci(2026)]{williams2026rltt}
Jonathan Williams and Esin Tureci.
\newblock Prioritize the process, not just the outcome: Rewarding latent
  thought trajectories improves reasoning in looped language models.
\newblock \emph{arXiv preprint arXiv:2602.10520}, 2026.

\bibitem[Zheng et~al.(2023)Zheng, Chiang, Sheng, et~al.]{zheng2023judge}
Lianmin Zheng, Wei-Lin Chiang, Ying Sheng, et~al.
\newblock Judging {LLM}-as-a-judge with {MT-Bench} and {Chatbot Arena}.
\newblock In \emph{Advances in Neural Information Processing Systems, Datasets
  and Benchmarks Track}, 2023.
\newblock arXiv:2306.05685.

\bibitem[Zhu et~al.(2025)Zhu, Wang, Hua, Zhang, Li, et~al.]{zhu2025ouro}
Rui-Jie Zhu, Zixuan Wang, Kai Hua, Tianyu Zhang, Ziniu Li, et~al.
\newblock Scaling latent reasoning via looped language models.
\newblock \emph{arXiv preprint arXiv:2510.25741}, 2025.

\end{thebibliography}

\appendix

\section{Corrections to versions 1 and 2}\label{app:corrections}

\begin{table}[h]
\centering
\footnotesize
\caption{Every headline claim of version~1, its status in version~2's erratum, and its value here.}
\label{tab:corrections}
\begin{tabular}{@{}>{\raggedright\arraybackslash}p{4.3cm}>{\raggedright\arraybackslash}p{1.9cm}>{\raggedright\arraybackslash}p{2.4cm}>{\raggedright\arraybackslash}p{3.1cm}>{\raggedright\arraybackslash}p{2.8cm}@{}}
\toprule
Claim & Version 1 & Version 2 & This version & Cause \\
\midrule
Evaluator accuracy & 95.2\% & 63.9\% & 64.7\% (Thinking, 1,024); 65.0\% (RLTT, 768) & position bias \\
Relational linear probe & 84.5\% (last loop) & 56.5\% (all loops) & 60.6\% (all loops); 61.5\% (last loop) & leakage (v1); truncation (v2) \\
Single-response linear probe, per reply & 21.75\% & 54.2\% (all loops) & 54.3\% (all loops); 54.9\% (last loop) & leakage \\
Relational minus single response & 63 points (pair vs per-reply accuracy) & $+2.3$ (pair vs per-reply); $+1.0$ ranking the same pairs (all loops) & ranking the same pairs: $+0.7$ (last loop, not distinguishable from zero), $+1.3$ (all loops) & as above \\
Test accuracy by epoch & 83.3 $\to$ 95.2 $\to$ 62.4\% & fixed-order, not corrected & 63.7 $\to$ 64.7 $\to$ 65.3\% & position bias \\
Learning-rate schedule preserved the peak & claimed & stands & withdrawn: there is no peak & position bias \\
Random-order training accuracy ``deflated'' & claimed & stands & random order balances the offset, the fixed-order test did not & position bias \\
Score correlation as the antisymmetry metric & claimed & corrected & order-independent accuracy & position bias \\
Rejected replies longer; no surface shortcut & 500 pairs & stands & longer reply preferred in 52.9\% & small sample \\
Evaluator exceeds reward models (72--75\%) & claimed & withdrawn & withdrawn & \\
Ouro is the only public looped model & claimed & not addressed & withdrawn (Huginn) & \\
\bottomrule
\end{tabular}
\end{table}

Version~1's architecture search, its training-time accuracies for the single-response evaluators, and its error analysis of the 95.2\% are not re-measured here and are left out.

\section{Reproducibility}\label{app:repro}

\paragraph{Fixed in advance versus after the fact} The pair-disjoint split and the probe definitions and regularization come from version~2's audit; the evaluator checkpoints are version~1's (version~2's audit evaluated only epoch~2). Truncating from the left, counting ties as one half, the length and norm baselines, and the per-epoch analysis were introduced for this version after the problems with versions 1 and~2 were found. An outside review of a draft of this version found that three of the four probe fits had stopped at an iteration cap, so their results depended on the number of CPU threads; the probes reported here are fit to convergence, and their results are identical with one and with eight threads. The nonlinear test of Section~\ref{sec:relational} was designed after the linear results were known; its protocol (data, validation slice, grid, five seeds, endpoint and verdict rule) was hashed before its first fit. Both networks read the same 512 principal components of the features, fit on training pairs only.

\paragraph{Models and data} Ouro-2.6B (revision \texttt{1ed04250}), Ouro-2.6B-Thinking (\texttt{f1edd81e}) and Ouro-RLTT (a research checkpoint that is not public; shard hashes in \texttt{v3/INPUTS\_OFFLINE.sha256}). HH-RLHF revision \texttt{09be8c5b}. The probes' 40,000 pairs and their split come from seed 20260711. Evaluator checkpoints are version~1's epochs 1--5 (epoch~2 SHA-256 \texttt{3630c209\ldots}); the Ouro-Thinking runs read state-only copies with identical weights. Probe feature extraction for both backbones and the Ouro-Thinking evaluator runs used an NVIDIA A40 (Runpod); the Ouro-RLTT evaluator runs used an RTX 5070 Ti laptop GPU. Scripts, results, per-pair outputs and fitted probe weights are in \url{https://github.com/VykosMolt/Hidden-State-Evaluator} under \texttt{v3/}, at commit \texttt{ae19e0e6c9216353ae85d6bb38dcb8aa2ac332d3}. Its \texttt{README.md} maps the numbers in this paper to the scripts and files that produce them (the five random-order accuracies of Section~\ref{sec:position} are the means of the fixed-order and swapped-order accuracies it lists; the capacity-matched check of Section~\ref{sec:relational} is in \texttt{results\_thinking\_last.json}), and \texttt{INPUTS\_OFFLINE.sha256} lists, with SHA-256 hashes, the inputs that are not in the repository: the feature shards, the evaluator checkpoints and the Ouro-RLTT weights.

\paragraph{Implementation notes} The probes and the evaluator use Ouro-RLTT's tokenizer for every backbone, as version~2's audit did; it differs from Ouro-Thinking's only in which tokens are marked special, which does not change how HH-RLHF text is encoded. Ouro's early-exit threshold changes only which loop's logits are used; every forward pass returns all four loop states, so the evaluator's training setting (threshold 0.87) does not change its inputs.

\paragraph{Run-to-run variation} Rerunning the evaluator on Ouro-RLTT with the settings of version~2's audit gives 64.5\% with ties counted as errors, against the audit's 63.9\%. Of the 8,552 pairs, 311 change sign between the two runs; their scores are close to zero (median magnitude 0.32, against 1.47 over all pairs). A plausible cause, which I did not test, is bfloat16 rounding that depends on how pairs are batched. The two runs differ by 0.55 points.

\end{document}